\documentclass{l4dc2024}

\title[Pixel SafeRL]{State-Wise Safe Reinforcement Learning with Pixel Observations}

\usepackage{times}
\usepackage{hyperref}
\usepackage{url}
\usepackage{amsmath}
\usepackage{wrapfig}

\usepackage{algorithm2e}

\usepackage{amsfonts}
\usepackage{algorithm}
\usepackage{color,soul}
\usepackage{graphicx}
\usepackage[font=small,labelfont=bf]{caption}

\newenvironment{myitemize}{\begin{list}{$\bullet$}
{\setlength{\topsep}{1mm}
\setlength{\itemsep}{0.25mm}
\setlength{\parsep}{0.25mm}
\setlength{\itemindent}{0mm}
\setlength{\partopsep}{0mm}
\setlength{\labelwidth}{15mm}
\setlength{\leftmargin}{4mm}}}{\end{list}}

\coltauthor{\Name{Simon Sinong Zhan} \Email{SinongZhan2028@u.northwestern.edu}\\
 \Name{Yixuan Wang} \Email{yixuanwang2024@u.northwestern.edu}\\
 \Name{Ruochen Jiao} \Email{RuochenJiao2024@u.northwestern.edu}\\
 \Name{Qi Zhu} \Email{qzhu@northwestern.edu}\\
 \addr Northwestern University, Evanston, IL 60208, USA}

\coltauthor{\Name{Qingyuan Wu} \Email{Qingyuan.Wu@liverpool.ac.uk}\and
 \Name{Qi Zhu} \Email{Chao.Huang2@liverpool.ac.uk}\\
 \addr The University Of Liverpool, Liverpool, L69 3BX, UK}

\author{%
 \Name{Simon Sinong Zhan} \Email{SinongZhan2028@u.northwestern.edu}\\
 \addr Northwestern University, Evanston, IL 60208, USA
 \AND
 \Name{Yixuan Wang} \Email{yixuanwang2024@u.northwestern.edu}\\
 \addr Northwestern University, Evanston, IL 60208, USA
 \AND
 \Name{Qingyuan Wu} \Email{Qingyuan.Wu@liverpool.ac.uk}\\
 \addr The University Of Liverpool, Liverpool, L69 3BX, UK
 \AND
 \Name{Ruochen Jiao} \Email{RuochenJiao2024@u.northwestern.edu}\\
 \addr Northwestern University, Evanston, IL 60208, USA
 \AND
 \Name{Chao Huang} \Email{thomashuangchao@gmail.com}\\
 \addr The University Of Southampton, Southampton, SO17 1BJ, UK
 \AND
 \Name{Qi Zhu} \Email{qzhu@northwestern.edu}\\
 \addr Northwestern University, Evanston, IL 60208, USA
}

\begin{document}

\maketitle

\begin{abstract}%
In the context of safe exploration, Reinforcement Learning (RL) has long grappled with the challenges of balancing the tradeoff between maximizing rewards and minimizing safety violations, particularly in complex environments with contact-rich or non-smooth dynamics, and when dealing with high-dimensional pixel observations. Furthermore, incorporating state-wise safety constraints in the exploration and learning process, where the agent must avoid unsafe regions without prior knowledge, adds another layer of complexity. In this paper, we propose a novel pixel-observation safe RL algorithm that efficiently encodes state-wise safety constraints with unknown hazard regions through a newly introduced latent barrier-like function learning mechanism. As a joint learning framework, our approach begins by constructing a latent dynamics model with low-dimensional latent spaces derived from pixel observations. We then build and learn a latent barrier-like function on top of the latent dynamics and conduct policy optimization simultaneously, thereby improving both safety and the total expected return. Experimental evaluations on the safety-gym benchmark suite demonstrate that our proposed method significantly reduces safety violations throughout the training process, and demonstrates faster safety convergence compared to existing methods while achieving competitive results in reward return. \href{https://github.com/SimonZhan-code/Step-Wise_SafeRL_Pixel}{Github Code and Demos.}
\end{abstract}


\section{Introduction}
Reinforcement Learning (RL) has demonstrated promising achievements in addressing control problems across diverse domains including robotics~\citep{zhao2020sim}, games~\citep{silver2016mastering}, and cyber-physical systems~\citep{liu2020parallel, yu2021review, wang2020energy}. Despite its potential, the occurrence of state-wise safety violations during the learning exploration phases has restrained industries from integrating RL methods into safety-critical applications such as traffic control~\citep{wei2018intellilight}, autonomous driving~\citep{kiran2021deep}, and power grid~\citep{duan2019deep}. 

Conventional safe RL methods are based on the Constrained Markov Decision Process (CMDP) paradigm~\citep{altman2021constrained}, which encodes the safety constraints through a cost function of safety violation and reduces the exploration space to where the trajectory-level discounted cumulative expected cost below a predefined threshold.  However, a fundamental issue arises from the soft nature of the safety constraints in CMDP, which can hardly capture and enforce stringent reachability-based state-wise safety constraints~\citep{wang2023enforcing}. On the other hand, the state-of-the-art theoretical control approaches such as barrier theory~\citep{ames2019control}, contraction theory~\citep{tsukamoto2021learning}, and reachability analysis~\citep{bansal2017hamilton, xue2023reach, wang2023polar} have their advantages in effectively encoding and optimizing state-wise safety. There are attempts to combine the aforementioned methods with model-based RL techniques~\citep{choi2020reinforcement, dawson2022safe, wang2023joint, wang2023enforcing} to train an RL policy with safety guarantee. Nonetheless, those control-theoretical RL methods encounter significant challenges and limitations when operating in unknown environments with pixel observations. Overall, we summarize the challenges with the existing approaches when dealing with pixel-observation state-wise safe RL problems as follows. 
\begin{myitemize}
    \item \textbf{Challenge 1:} The existing CMDP problem is too soft to encode the state-wise safety constraint. 
    \item \textbf{Challenge 2:} Control-theoretical approaches typically rely on relatively low-dimensional state spaces with clear physical interpretations, making it challenging for them to scale and adapt to the complexities posed by high-dimensional pixel observations.
    \item \textbf{Challenge 3:} Control-theoretical approaches typically rely on prior knowledge of the unsafe regions and require explicit models of the environment dynamics which are often smooth (ODE, SDE, etc.). However, such requirements become impractical in the context of unknown contact-rich and non-smooth environments, with no prior knowledge about the hazard regions, a common scenario in RL setups. 
\end{myitemize}
To address these challenges, we introduce a novel state-wise safety-constrained RL framework tailored for image observation in unknown environments. Our framework aims to optimize control policies and minimize safety violations during the training and exploration stages. Specifically, for \textbf{Challenge 2}, we learn to compress the pixel observation into a low-dimensional latent space. For \textbf{Challenge 3}, with the compressed latent space, we further learn the latent MDP-like dynamics in it to deal with the data from contact-rich and non-smooth unknown environments.  We establish a latent barrier-like function on it to encode the state-wise safety constraint for \textbf{Challenge 1}.

Our framework jointly conducts the latent modeling, latent barrier-like function learning, and policy optimization in an actor-critic framework, leading to improvements in both safety and overall return. Compared to model-free approaches, our approach is sample-efficient by generating training data from the latent model to avoid unsafe interactions with environmental hazards. In comparison to existing model-based methods, our approach directly enforces safety by the power of the barrier function, resulting in fewer state-wise safety violations during the training process and faster convergence of cost return, while achieving similar total reward return.

The paper is organized as follows. Section \ref{related_work} introduces related works; Section \ref{formulation} elaborates on the formulation of our problem; Section \ref{approach} presents our joint learning framework, including latent modeling, latent barrier-like function learning, and policy optimization; Section \ref{experiment} showcases experiments setup and results; Section \ref{conclusion} concludes our paper.

\section{Related Works}
\label{related_work}

\textbf{Safe RL by CMDP with High-Dimensional Input:} The primal-dual approaches have been widely adopted to solve the Lagrangian problem of CMDP in a model-free manner, such as PDO~\citep{chow2017risk}, OPDOP~\citep{ding2021provably}, CPPO~\citep{stooke2020responsive},  FOCOPS~\citep{zhang2020first}, CRPO~\citep{xu2021crpo}, and P3O~\citep{shen2022penalized}, which typically deal with state input rather than RGB pixels as in our work. The RL community has recently shown a growing interest in the challenge of learning policies from rich, high-dimensional data inputs~\citep{rafailov2021offline, zhang2020learning, hafner2019learning, hafner2020mastering}. Recent efforts have been made to address CMDP with pixel observations~\citep{yarats2021improving}. However, the model-free CMDP approaches are sample-inefficient in dealing with the high dimensional image input~\citep{shang2021reinforcement} and thus lead to a large number of safety violations during the exploration. Recent approaches have aimed to mitigate these issues by mapping image observations to low-dimensional spaces to reduce sampling complexity and number of safety violations~\citep{as2022constrained, hogewind2022safe} in a model-based manner for CMDP. Despite this, CMDP approaches still have difficulty enforcing the state-wise safety constraints~\citep{xiong2023provably, wang2023enforcing}.

\textbf{Safe RL with Sate-wise Constraint and High-Dimensional Input:} There are some efforts to combine classic control theory methods to enforce safety in safe RL, such as reach-avoid RL~\citep{hsu2021safety}, contraction RL~\citep{sun2021learning}. Among them, barrier function is one of the most powerful approaches. The barrier function is a formal certificate that is affiliated with a control policy to prove state-wise safety of a dynamical system~\citep{ames2016control, dawson2023safe}. In classical control theory, a common approach is to relax the conditions of the barrier function into optimization formulations such as linear programs~\citep{wang2023joint, yang2016linear} and quadratic programs~\citep{ames2019control, choi2020reinforcement}.  Recent work~\citep{wang2023enforcing, wang2023joint, cheng2019end, qin2021learning, dawson2022safe} proposes to jointly learn control policy and neural barrier function to optimize state-wise safety constraints in RL. However, a major problem of these approaches is their limited scalability to a higher-dimensional system let alone pixel observation. Efforts to encode state-wise safety using barrier functions derived from visual inputs have been made in~\citep{dawson2022learning, tong2023enforcing, cui2022learning}. However, these methods typically rely on a known dynamic model or require depth/distance information from the perception beyond the scope of unknown environments with RGB pixel observations as assumed in our work. Additionally, there have been attempts to transform high-dimensional state spaces into lower-dimensional representations and utilize sampling-based in-distribution barrier functions~\citep{castaneda2023distribution}. Nonetheless, this approach is limited to a perfect continuous dynamic model and necessitates the presence of an existing reference controller.

\section{Problem Formulation}
\label{formulation}
We consider an RGB-pixel-observation safe RL problem with an unknown environment. Assume the environment can be modeled as a finite-horizon Markov Decision Process (MDP) $\mathcal{M}\sim(\mathcal{S, O, A, P}, r, \gamma)$. $\mathcal{S} \subset \mathbb{R}^n (n \in \mathbb{Z}_+)$ stands for a continuous state space, $\mathcal{A} \subset \mathbb{R}^m(m \in \mathbb{Z}_+)$ stands for a continuous action space, and $s_{t+1}\sim \mathcal{P}(\cdot|s_t, a_t)$ stands for the unknown transition dynamic function of the environment, where $s_t\in\mathcal{S}$ and $a_t\in\mathcal{A}$. $\mathcal{O} \subset \mathbb{R}^{c_o \times h_o \times w_o}(c_o, h_o, w_o \in \mathbb{Z}_+)$ is the observation space captured by the camera module on the agent, $r: \mathcal{S} \times \mathcal{A}\times\mathcal{S}\rightarrow\mathbb{R}$ is the reward function of RL, and $\gamma \in [0, 1]$ is a discount factor. The observation space is dependent on the state space as there exists an unknown function that maps an underlying $s_t \in \mathcal{S}$ to the $o_t \in \mathcal{O}$ captured by the agent. We assume the control policy $\pi_{\theta}$ by the agent generates control actions $a_t$ by consuming the underlying state $o_t$, i.e., $a_t \in \mathcal{A} \sim \pi(\cdot|o_t)$. It is a realistic assumption as in the real world, the agent usually is not able to get the ground truth of its state and takes actions based on the observed image. We consider state-wise safety for this MDP formulation. Assume there exists some \textbf{unknown} unsafe spaces as a set $S_u \subset \mathbb{R}^n$, the state-wise safety violation is defined as $s_t \in S_u$, \textit{assuming there exists a safety detector $\kappa$ that can check the safety violation. }Overall, the state-wise safe RL problem with pixel observation is defined as
\begin{equation}
    \begin{aligned}
        &\max_{\theta} J(\pi_\theta) =\mathbb{E}_{\mathcal{P}(\cdot|s_t, a_t)  }\left[\sum_{{t}=0}^{T} \gamma^{t} r(s_t, a_t, s_{t+1})\right],\text{s.t.}, \sum_{t=0}^{T} \kappa(s_t) \leq D, a_t \sim \pi_{\theta}(\cdot|o_t).
    \end{aligned}
\end{equation}
where $\kappa(s_t) \in \{0, 1\}$ indicates safety violation, $D \in \mathbb{R}$ is a safety violation budget. In real-life safety-critical systems, we would like the safety violations as few as possible, therefore we would like to minimize the number of safety violations towards zero during learning, e.g., $D \rightarrow 0$.

\section{Our Approach}
\label{approach}
\begin{figure}[h]
    \centering
    \includegraphics[width=0.9\linewidth]{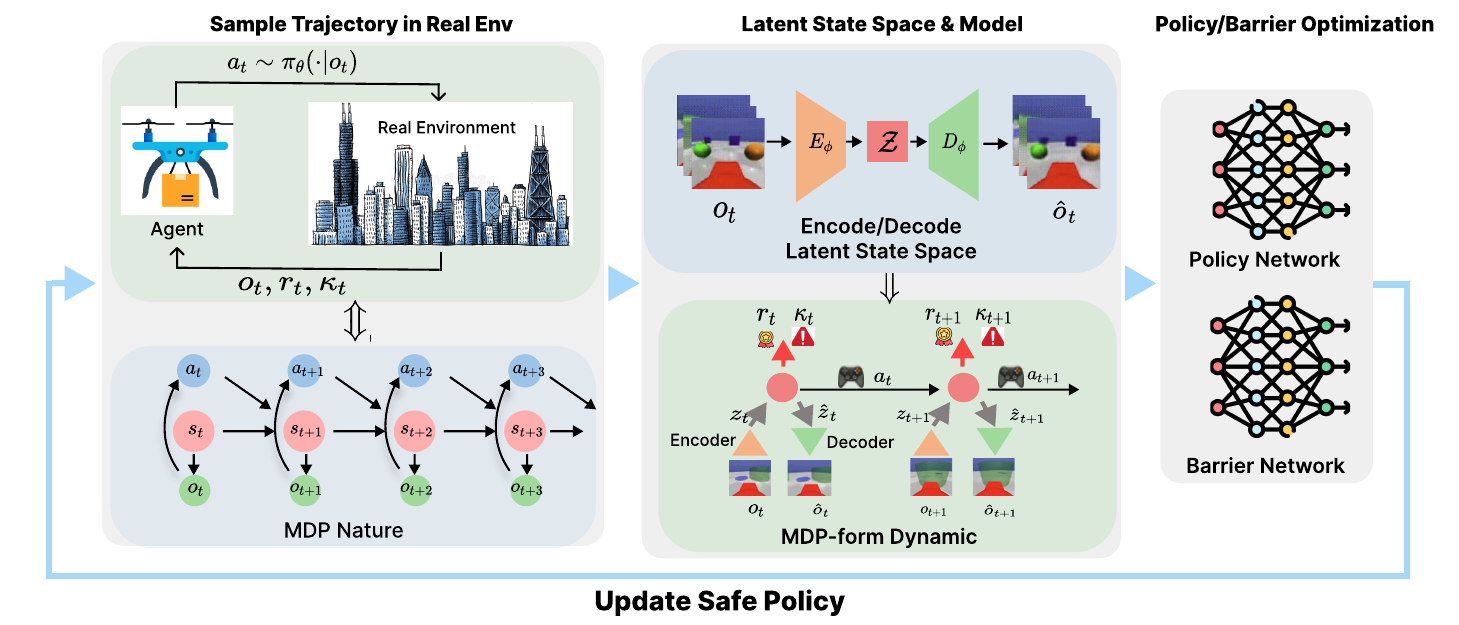}
    \caption{Overview logistic diagram of our joint learning framework. We sample pixel-observation data from real environments,  learn to compress the image data to a low-dimensional latent model with MDP-like latent dynamics, and then learn a latent barrier-like function on top of it to encode state-wise safety constraints and conduct policy optimization for safe exploration and performance improvement simultaneously. 
    % $\theta$ stands for the parameters of the policy network. $\phi$ stands for the parameters of the latent model.}
    }
    \label{fig:models}
\end{figure}

We introduce our joint learning framework for state-wise safe RL with pixel observations. Figure~\ref{fig:models} shows the high-level overview of our approach including latent modeling, barrier-like function learning, and policy optimization. To alleviate the complexity of the high-dimensional pixel observations (\textbf{Challenge 2}), we first design our approach to learn to compress the image observation into a low-dimensional latent vector by reconstructing it in a VAE-like manner and further build forward dynamics within this latent space, i.e., latent dynamics. Due to the power of learning, such a latent modeling approach can consume the data from the non-smooth contact-rich environment observations and is able to learn the unsafe regions by our design of latent safety predictor (\textbf{Challenge 3}). Hence, the MDP-like latent model functions as a generative model for producing synthetic data used in training. Consequently, our approach operates in a model-based fashion, minimizing interactions with the actual environment and thereby reducing safety violations. With the foundation of latent dynamics, we then build a latent barrier-like function on top of the latent model to encode state-wise safety constraints to further improve safety (\textbf{Challenge 1}) by training from the synthetic data with the safety labels from a learned latent safety predictor. It is worth noting that the training gradient from the barrier-like function can back-propagate to the control policy for safer actions. Meanwhile, we conduct policy optimization to improve the total expected return in a model-based manner. The overview of our approach is in Algorithm~\ref{overall_algo}. Next, we are going to introduce each component by each subsection.
\begin{algorithm}
\caption{State-wise Safe RL with Pixel Observations}
\label{overall_algo}
% \SetLine
\KwData{Unknown environment with an initial policy $\pi_{\theta}$, generated horizon $H$, action repeat $R$, collect interval $C$, batch size $B$, chunk length $L$, total episodes $E$, episode length $T$}
\KwResult{Policy $\pi_{\theta}$ with barrier-like function $B_{\theta}$ and the latent model with $\phi$}

Initialize dataset $\mathcal{D}$ with random seed episodes, models with parameters $\theta$, $\phi$, and $\omega$.\;

\For{\texttt{epoch} \textbf{in} $E$}{
    \For{\texttt{update step} \textbf{in} $C$}{
        Sample batch of sequence chunks $\{(o_t, a_t, r_t, \kappa_t)_{t=k}^{L+k}\}_{i=1}^B\sim\mathcal{D}$.\;
        
        Train \textit{latent model \ref{latent model}} and calculate $\mathcal{L}_{m}$ from \textit{Equation~\eqref{latent_loss}}.\;
        
        Update $\phi\leftarrow\phi+\varphi\nabla_{\phi}\mathcal{L}_{m}$.\;\hfill\tcp{Update the latent model in one pass}
        Generate trajectories $\{(z_t, a_t, \hat{r}_t, \hat{\kappa}(z_t))_{t=\tau}^{\tau+H}\}_{i=1}^{B\times(L+k)}$ using current policy in latent space.\;
        
        Compute $\mathcal{L}_{b}$ from \textit{Equation~\eqref{l_barrier}},
         $\mathcal{L}_{p}$ from \textit{Equation~\eqref{l_policy}}.\;
         
        Update $\theta\leftarrow \theta+\alpha\nabla_{\theta}\mathcal{L}_{p}$.\;\hfill
        \tcp{Update barrier and policy in one pass}
        Update $\omega\leftarrow \omega+\alpha\nabla_{\omega}\sum\frac{1}{2}\lVert v_{\omega}(z_t)- \widehat{V}_\omega^\pi(z_t)\lVert$.\;\hfill
        \tcp{Update value network}
    }
    \For{\texttt{i} in $\frac{T}{R}$}{
        Compute $z_t$ and $a_t$ from \textit{latent model} and $\pi_{\theta}$,
        add exploration noise on top $a_t$.\;
        
        $r_t, \kappa_{t+1}, o_{t+1}\leftarrow$\texttt{env.step($a_t$)}.\;\hfill
        \tcp{Deploy in real env to collect traj}
    }
    Add the new trajectory to $\mathcal{D}$.\;
}
\end{algorithm}

\subsection{Pixels to Latent State Space with Latent Dynamics}
\label{latent model}
Our framework first learns to transform the environment MDP (as in Section~\ref{formulation}) into an MDP-like latent model with a low-dimensional latent space $(\mathcal{Z}, \mathcal{A}, \mathcal{T}, \hat{r}, \hat{\kappa}, \gamma)$. The comprehensive depiction of our latent space is illustrated in Figure~\ref{fig:models}. To streamline computational complexity, we utilize a VAE-like Encoder($E_\phi$)-Decoder($D_\phi$) structure to encode pixel observations $\mathcal{O}$ into low-dimensional \textbf{latent state spaces} $\mathcal{Z}$ ($\mathcal{Z} \sim E_\phi(\mathcal{O})$, $\mathcal{O} \sim D_\phi(\mathcal{Z})$). In addition, we construct a \textbf{reward predictor} and \textbf{safety predictor} model for predicting the reward $\hat{r}_t(z_t, a_t)$ and safety status $\hat{\kappa}_t(z_t) \in \{0, 1\}$ associated with the respective latent state $z_t$ for a given $o_t$. It is important to note that in our formulated problem (see Formulation~\ref{formulation}), we presume the existence of a safety detector $\kappa(s_t)$, which might be a fusion of different types of sensors. The role of the safety predictor $\hat{\kappa}(z_t)$ is to predict the output of this safety detector $\kappa(s_t)$, such that the latent space can tell the safety violation within it. Furthermore, to emulate the dynamics of the MDP in Section~\ref{formulation}, an inference \textbf{transition model} $\mathcal{T}$ is applied to the latent state space. This model, taking a latent state $z_t$ and an action from the policy $a_t$ as input, outputs the Gaussian distribution of the $z_{t+1}$, i.e., $z_{t+1} \sim \mathcal{T}(\cdot | z_t, a_t)$. We note that this latent model shares the same control policy with the real environment MDP.
With these components, we can fully capture the dynamical nature of the environment in the latent space (i.e., $s_{t+1} \sim \mathcal{P}(\cdot| s_t, a_t) \rightarrow z_{t+1} \sim \mathcal{T}(\cdot | z_t, a_t)$) with reward and safety signals. Besides, this latent model can serve as a generative model to synthesize data for training the control policy, i.e., we can sample latent trajectory data $\{(z_t, a_t, \hat{r}_t, \hat{\kappa}_t)\}_{t=0}^{T}$ and thus reduce the agent's interactions with the real environment during the training to avoid unsafe manners. We train our latent model by using trajectories chunk with time length $T$ from the data buffer of the real environment MDP, and we define the loss function as follows. 
\begin{equation}
\label{latent_loss}
    \mathcal{L}_{m} = \sum_{t=0}^{T} \left(\text{KL}(z_t||E_\phi(o_t)) + \text{MSE}(\hat{r}_t, r_t) + \text{MSE}(\hat{\kappa}_t, \kappa_t) + \text{MSE}(\hat{o}_t, o_t) \right)
\end{equation}
The first \textit{KL-Divergence Loss}  measures the distribution difference between inferred latent state $z_t$ and ground-truth compressed from real observation $E_\phi(o_t)$, which is used to update our transition model $\mathcal{T}$; The second and third \textit{MSE Loss} serve to learn reward predictor and safety predictor; And the last \textit{MSE Loss} captures loss of observation compression process through reconstruction from latent space. All the components of the latent model share common parameters $\phi$ and therefore are updated at the same backward pass. This low-dimensional latent model with assumptions on the MDP nature of dynamics can learn the non-smooth transition to address the aforementioned \textbf{Challenge 2, 3}. For implementation details, the probabilistic transition model $\mathcal{T}$ is implemented as an RNN, The observation encoder and decoder are CNN and transposed CNN respectively, and the reward predictor and safety predictor are all constructed as DNNs. We build our latent model on top of existing RSSM structure~\citep{hafner2019learning}. It is worth noting that we leverage a different notion of latent state space and learn the safety predictor of the environment.

\subsection{Latent Barrier-like Function Learning}
\label{barrier}
Based on the previous latent model, we introduce our barrier-like function on latent state space to intuitively enforce the forward invariance for state-wise safety constraints where the safe and unsafe latent states can be separated from the aforementioned safety predictor. 
\begin{definition}
    Given a policy $\pi_{\theta}$, $B_{\theta}$ is a barrier-like function of the latent state space if it satisfies the conditions below:
\begin{equation} \label{eq:barrier}
    \begin{aligned}
     B_\theta(z_{s}) > 0, \quad 
    B_\theta(\mathbb{E}(z_t|z_{t-1})) - B_\theta(z_{t-1}) + \alpha B_\theta(\mathbb{E}(z_t|z_{t-1})) > 0, \quad
    B_\theta(z_{u}) < 0 
    \end{aligned}
\end{equation} 
where $z_t \sim \mathcal{T}(\cdot | z_{t-1}, \pi_{\theta}(z_{t-1})), \theta$ stands for the parameters of the barrier-like function and policy network. $z_s \in \mathcal{Z}_s \subset \mathcal{Z}$ stands for state in safe latent state set $\mathcal{Z}_s$, $z_u \in \mathcal{Z}_u \subset \mathcal{Z}$ stands for state in unsafe latent state set $\mathcal{Z}_u$, and $\alpha$ is a class-$\mathcal{K}$ function. 
\end{definition}
Note that $\mathcal{Z}_s$ and $\mathcal{Z}_u$ are categorized by safety predictor learned in the latent model~\ref{latent model}, i.e.,  $\hat{\kappa}({z}_t)=1, z_t\in\mathcal{Z}_u$, otherwise $ z_t\in\mathcal{Z}_s$. The latent barrier-like function offers a state-wise definition of safety.  The idea is to have the agent start in the $z_t \in \mathcal{Z}_s, B(z_t) > 0$ and, by encoding the positivity of the time derivative (as approximated by the second condition in Equation~\eqref{eq:barrier}), ensure that the expected subsequent state maintain this positivity, i.e., $B_{\theta}(\mathbb{E}[z_{t+1}]) > 0$. This approach results in the invariance of the agent within the safe state set in expectation. However, due to partial observability, there may be instances where the agent unintentionally enters the unsafe state set. When this occurs, the barrier-like function yields a negative value. Still, the positivity of the consecutive state function value difference guides the agent away from unsafe regions and towards states characterized by a positive barrier value. In contrast, CMDP approaches primarily focus on minimizing the total cost over an entire trajectory in expectation without considering this state-wise encoding of safety. 
\begin{remark}
\label{z_t_disclaimer}
    In this study, we add a stochastic element to the mean of the distribution derived from the transition model $\mathcal{T}$ and denote it as $z_t$ for subsequent computations. This approach is commonly employed in various model-based methods~\citep{as2022constrained, hogewind2022safe}. Our experimental findings reveal that without incorporating stochasticity into the mean significantly deteriorates efficacy of the reconstruction process due to the deterministic shortcut from encoder output directly to decoder input~\citep{hafner2019learning}. This, consequently, results in poor overall quality of latent model learning and policy performance. 
%Achieving invariance under time-varying distributions represents a longstanding and formidable challenge within the Stochastic Optimal Control. Previous efforts have attempted to address this challenge by leveraging various martingale theories~\citep{prajna2004stochastic, feng2020unbounded, guo2023ito}, albeit within the context of SDE dynamics. Furthermore, alternative approaches have been explored, such as computing the probabilistic control invariance set within a MDP framework~\citep{gao2020computing}, and some sampling approaches~\citep{wang2023enforcing}. However, these methods are notably time-consuming and are not compatible with our Pixel RL setting.
\end{remark}
We implement the latent barrier-like function as a dense neural network and derive the following loss vector with inspiration from~\citep{qin2021learning}.
\begin{multline}
\label{l_barrier}
    \mathcal{L}_{b} = [\text{ReLU}(\eta - B_\theta(z_{s})),
    \text{ReLU}(\eta - (B_\theta(z_t) - B_\theta(z_{t-1}) + \alpha (B_\theta(z_t)))), \text{ReLU}(\eta + B_\theta(z_{u}))]
\end{multline}
The first term penalizes non-positive safe states; the second term enforces the positivity of the approximated time derivative condition; the third term penalizes the non-negative unsafe states of the function. We apply a small positive learning margin $\eta \in \mathbb{R}$ to make the optimization more feasible. Note that since the problem possesses partial observability, the above formulation can only enforce the forward invariance without a formal guarantee.

\subsection{Policy Optimization}
To optimize the total rewards while considering state-wise safety, we formulate an actor-critic approach with barrier-like function learning in the loop \textit{within the latent model}, by using the trajectories $\{z_t, a_t, \hat{r}_t, \hat{\kappa}_t\}$ generated by the latent model. With the encoder network embedded inside, the policy DNN $\pi_{\theta}(\cdot|E_\phi(o_t))$ (or equivalently $\pi_{\theta}(\cdot|z_t), z_t \approx E_\phi(o_t)$) outputs action $a_t$ in a Gaussian distribution, which is randomly sampled for training and provides mean action value for evaluation.  
% The policy model is implemented as a dense neural network with parameter $\theta$. We form the policy network output into a Gaussian distribution. The action, $a_t$, is randomly sampled from the distribution for training and sampled mean for evaluation. 
The value (critic) function of RL $v_{\omega}(z_t)$ can be expressed as $v_{\omega}(z_t) = \mathbb{E}_{\pi_{\theta}(\cdot|z_t)}\left[\sum_{t=\tau}^{\tau+T}\gamma^t \hat{r}_{t}\right]$. We define the total expected return as $J_\phi(\pi_{\theta}) = v_{\omega}(z_0)$ and aim to reduce the following loss function for the overall RL objective. 
\begin{equation}
\label{l_policy}
\mathcal{L}_{p} = - \max J_\phi (\pi_{\theta}) + \beta^T \mathcal{L}_{b}
\end{equation}
where we add barrier loss function as a regularization term for safety. $\beta$ here is a coefficient vector corresponding to each term in Equation~\eqref{l_policy}. And we backward $\mathcal{L}_{p}$ through stochastic back-propagation to update the policy network and barrier-like function in the same pass. Specifically, for the critic network, we use the sampled synthetic latent trajectory $\{z_t, a_t, \hat{r}_t, \hat{\kappa}_t\}$ and Monte Carlo approach to provide the learning target $\widehat{V}_\omega^\pi(z_t) = \frac{1}{N}\sum_{i=1}^{N}\left[ \sum_{\tau=0}^{T}\gamma^{\tau}\hat{r}_\tau +  \gamma^{t+T} v_{\omega}(z_{t+T}) \right]$ for it to reduce the loss $\left \lVert v_\omega - \widehat{V} \right \rVert^2$. For the actor network, the policy gradient from $J_\phi$ is well-established as in~\citep{sutton2018reinforcement}.

% To update and back-propagate the policy network, we need to maximize the reward for a given trajectory, $J_\phi(\pi_{\theta}) = \frac{1}{T}\sum_{t=\tau}^{\tau+T} \widehat{V}_\omega^\pi(z_t)$ (where $\widehat{V}_\omega^\pi(z_t)$ is the value estimation on the latent model generated trajectories, we use the bootstrapping trick on the classic Monte Carlo estimation, which balances the bias and variance, and back-propagate the value network through supervised regression~\citep{hafner2020mastering}), and minimize the barrier function loss~\eqref{l_barrier}. Thus, we formulate the loss of policy network below.

% $\beta$ here is a coefficient vector corresponding to each term in Equation~\eqref{l_policy}. And we backward $\mathcal{L}_{p}$ through stochastic back-propagation to update the policy network and barrier function in the same pass.  

% \begin{remark}
%     With the effort of the latent model capturing the environment MDP, the policy and barrier function takes the latent state $z_t$, as input for both training and evaluation. For final online deployment, the pixel observation, $o_t$, first goes through the latent encoding to get corresponding $z_t$ (i.e., ) and then passes into the latent policy for further actions. 
% \end{remark}

% \vspace{+12pt}
\newpage
\begin{figure}[h]
\centering
\includegraphics[width=0.85\linewidth]{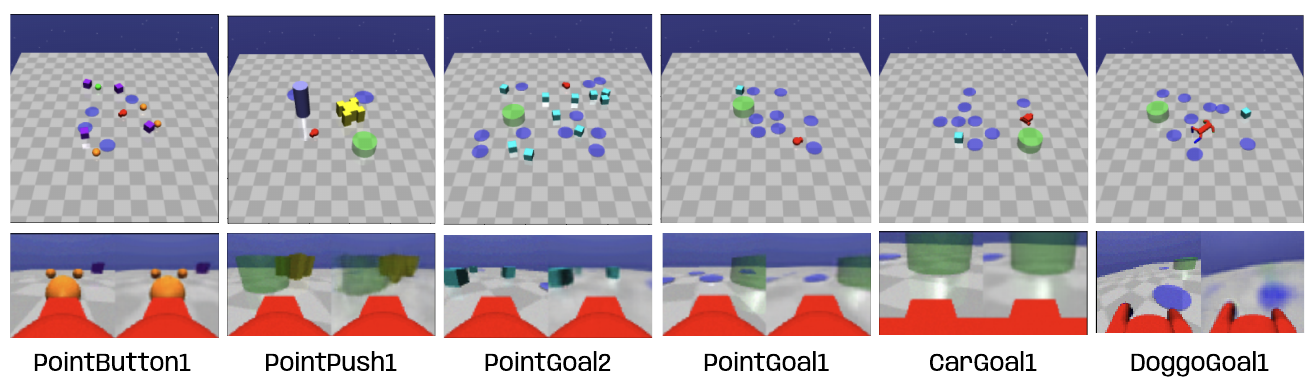}
\caption{This is a graphic expression of \texttt{Safety Gym SG6}~\citep{Ray2019} environments. Figures at the top row are with birdeye views of each benchmark. All images on the left-hand side in the bottom row are pixel observations taken by the agents, and all on the right-hand side are reconstructed images from our learned latent model, which appear similarly to the left observations.}
\label{fig:experiemnt_setup}
\centering
\end{figure}
\vspace{-8pt}
\section{Experiments}
\label{experiment}

\begin{wrapfigure}{i}{0.5\textwidth}
  \centering
  \includegraphics[width=0.5\textwidth]{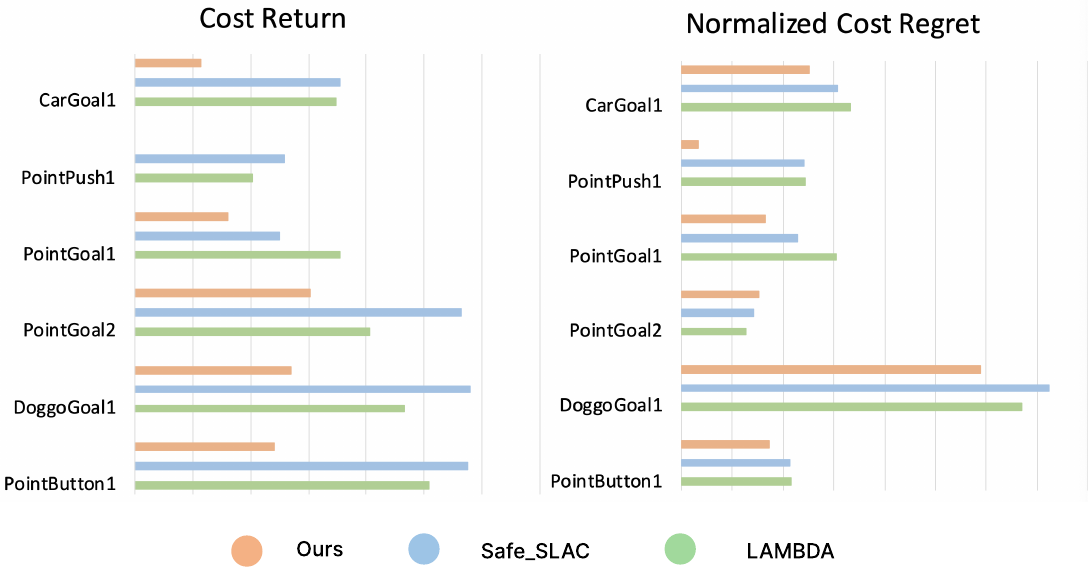}
  \caption{Left shows the \textit{Cost Return} of LAMBDA (green), Safe-SLAC (blue), and our approach (red) on the \texttt{Safety Gym} benchmarks;  Right illustrates the \textit{Normalized Cost Regret} with respect to PPO's. Overall, we can tell that our approach achieves fewer safety violations.}
  \label{fig:safety}
\end{wrapfigure}
We adopt the widely used \texttt{Safety Gym} \texttt{SG6} tasks~\citep{Ray2019} as test examples. For our approach and baselines, the agents take in $3\times 64 \times 64$ pixels images from the agents' point of view, as shown in Figure~\ref{fig:experiemnt_setup}. To the best of our knowledge, there is currently no state-wise safe RL approach capable of handling non-smooth environments encountered in the \texttt{Safety Gym} framework with pixel observations. Therefore, we compare to popular model-free CMDP methods including CPO, PPO-L, and TRPO-L, and pay more focus on LAMBDA~\citep{as2022constrained} and Safe-SLAC~\citep{hogewind2022safe}, two state-of-the-art model-based CMDP approaches that study safe RL with pixel observations. We train different algorithms with one million environment steps, except the \texttt{DoggoGoal1} with two million steps. We measure the performance of all approaches on the \textit{Cost Return} and \textit{Cost Regret}~\citep{Ray2019}, corresponding to state-wise safety in evaluation and training. In addition, we showcase the learning curve of our method compared with others' converged values after training.   
\begin{myitemize}
    \item The average \textit{Cost Return} for $N$ episodes is defined as $\hat{J}(\pi) = \frac{\sum_{i=0}^{N}\sum_{t=0}^{T_{ep}}c_t}{N}$, where $T_{ep}$ is the length of a single episode. Since $c_t$ is fixed for each step of safety violation in the \texttt{Safety Gym} environment, we can interpret this metric as the average safety violations performance of policy $\pi$ after each training episode.   
    \label{cost_return}
    \item The \textit{Cost Regret} is the sum of costs during the whole training process over the total interaction steps $T$. We defined as $\rho_c(\pi)=\frac{\sum_{t=0}^Tc_t}{T}$, $T$ is the total environment interactions, and $c_t$ is the cost corresponding to each interaction. This metric represents total safety violations accumulated in the entire training process.  
    \label{cost_regret}
    \item The average \textit{Reward Return} for $N$ episodes is defines as $\hat{R}(\pi)=\frac{\sum_{i=0}^{T_{ep}}r_t}{N}$, where $r_t$ is the reward received at time step $t$.  
\end{myitemize}

\begin{figure}[h]
    \centering
    \includegraphics[width=0.75\linewidth]{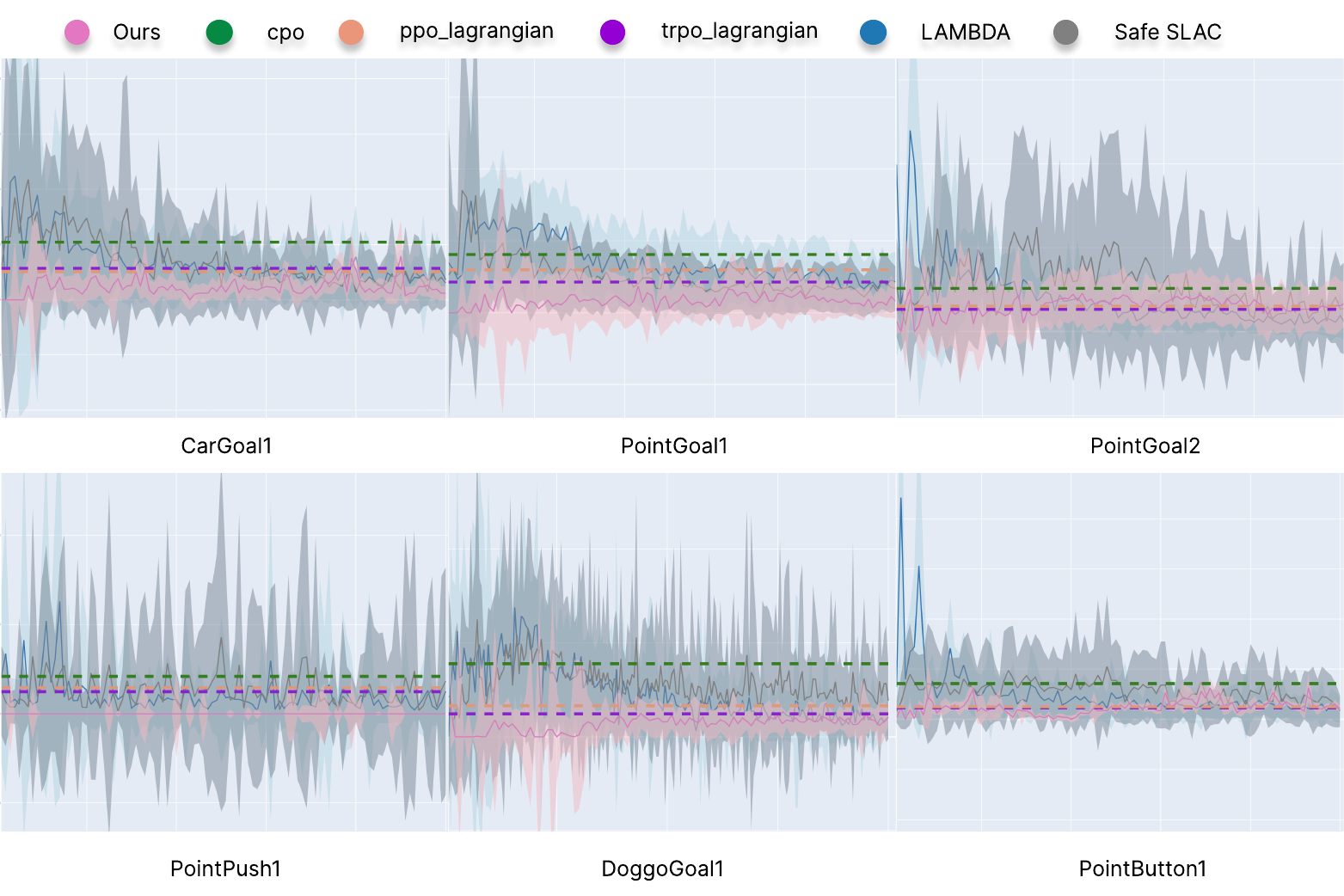}
    \caption{Dash-lines are \textit{Cost Return} of the model-free CPO, PPO-L, TRPO-L trained for 10M steps except \texttt{DoggoGoal1} with 50M training steps. Our approach shows faster cost convergence in all benchmarks compared with other approaches. Besides, for the majority of the benchmark, our approach can achieve the lowest converged cost return.}
    \label{fig:cost_return}
\end{figure}
% \vspace{-8pt}
\subsection{Safety Evaluation during Learning Explorations}
% \begin{figure}[h]
%     \centering
% \includegraphics[width=0.7\linewidth]{fig/safety_graph.pdf}
%     \caption{Left shows the \textit{Cost Return} of LAMBDA (green), Safe-SLAC (blue), and our approach (red) on the \texttt{Safety Gym} benchmarks;  Right illustrates the \textit{Normalized Cost Regret} with respect to PPO's. Overall, we can tell that our approach achieves fewer safety violations. 
%     % To achieve better return performance, we add extra stochastic noise in our policy. Though, intuitively, this stochastic noise leads to more safety violations in training process, our approach still has less cost regret(total safety violations in training) in majority of the benchmarks. 
%     }
%     \label{fig:safety}
% \end{figure}

Figure~\ref{fig:safety} shows \textit{Cost Return} and normalized \textit{Cost Regret} of our approach and baselines, where we normalize the \textit{Cost Regret} by dividing it by the \textit{Cost Regret} achieved by the \textbf{PPO} method, i.e., $\hat{\rho}_{ours}=\frac{\rho_{ours}}{\rho_{ppo}}$. It is worth noting that model-free approaches in principle lead to more safety violations~\citep{as2022constrained} and thus we mainly focus on model-based Safe-SLAC and LAMBDA for a fair comparison in Figure~\ref{fig:safety}. Compared to these two baselines, our approach exhibits a notable enhancement for \textit{Cost Return} in all benchmarks and consistently demonstrates lower \textit{Cost Regret}, except the \texttt{PointGoal2} environment. The results affirm the advantages of our latent barrier-like function learning for encoding state-wise safety constraints over the CMDP formulation in the baselines. 
In addition, from Figure \ref{fig:cost_return}, we can tell that our approach shows faster convergence in the \textit{Cost Return}. The reason is that during the training, our latent model quickly identifies and captures the majority of unsafe latent states by supervised learning. With more interactions, the latent barrier-like encoding hard state-wise safety constraint progressively forces the agent to take safer actions, leading to significantly lower \textit{Cost Return} compared to CMDP approaches.

\begin{remark}
\label{safety_explanation}
    Safety violations are unavoidable in our setup as the agent only receives partial image observations from a single front-view camera in an unknown environment without state information, where zero violation fundamentally is a hard problem to solve. In addition, due to learning errors, our latent model may not always accurately differentiate the safe and unsafe images, which could lead to safety violations. 
\end{remark}

% \begin{figure}[h]
%     \centering
%     \includegraphics[width=0.75\linewidth]{fig/cost_return.pdf}
%     \caption{Dash-lines are \textit{Cost Return} of the model-free CPO, PPO-L, TRPO-L trained for 10M steps except \texttt{DoggoGoal1} with 50M training steps. Our approach shows faster cost convergence in all benchmarks compared with other approaches. Besides, for the majority of the benchmark, our approach can achieve the lowest converged cost return.}
%     \label{fig:cost_return}
% \end{figure}

\vspace{-12pt}
\begin{figure}[h]
    \centering
    \includegraphics[width=0.7\linewidth]{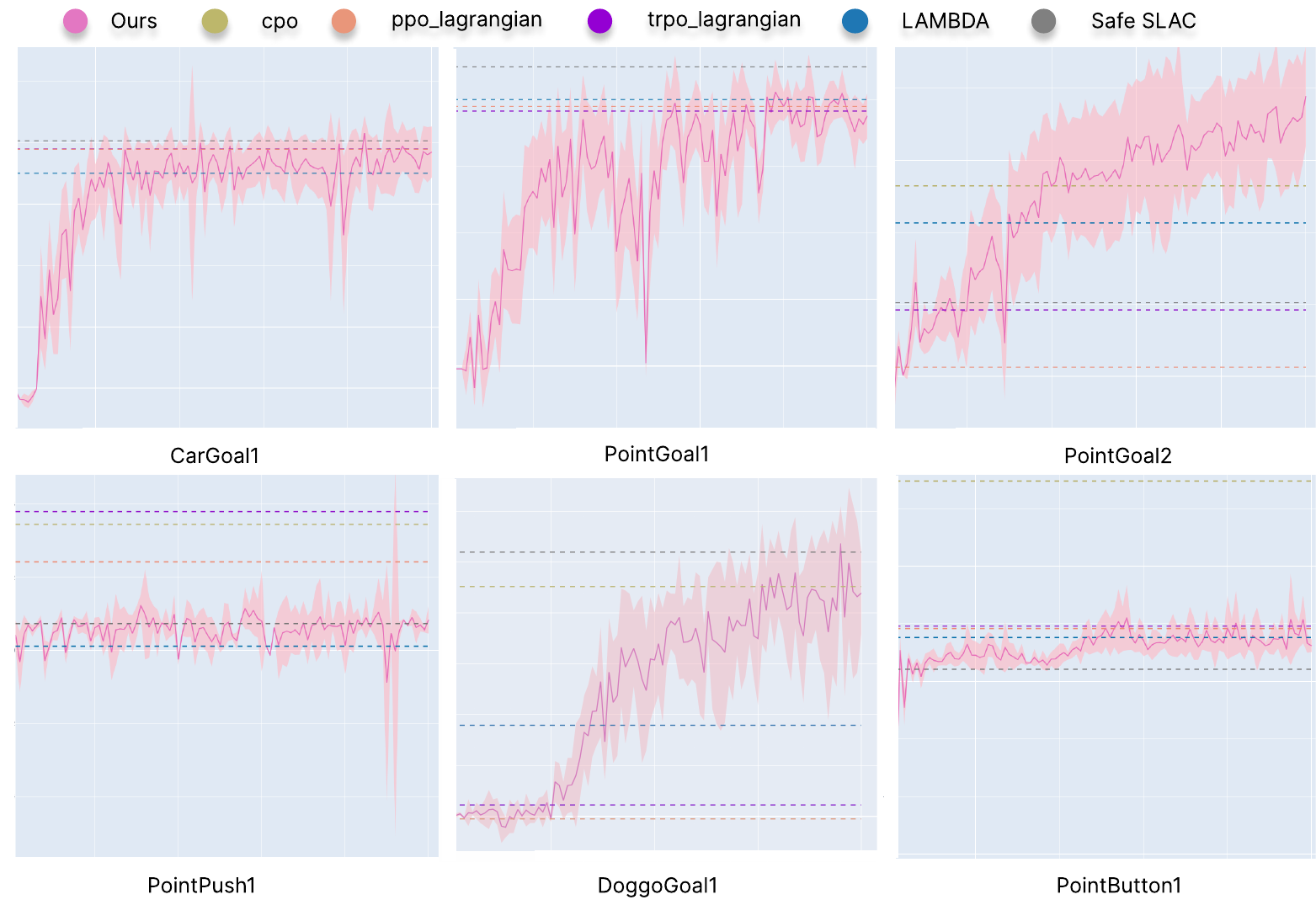}
    \caption{\textit{Reward Return}: The model-free methods CPO, PPO-L, and TRPO-L are trained for 10M steps except on \texttt{DoggoGoal1} with 50M steps. The model-based Safe-SLAC and LAMBDA are trained with 1M steps except on \texttt{DoggoGoal1} with 2M environment steps. For all the \texttt{SG6} benchmarks, our approach can achieve or even exceed other RL methods' performance at the end of training trials. }
    \label{fig:reward_return}
\end{figure}
\subsection{Reward Performance of the Learned Policies}
In principle, model-free approaches could obtain higher total expected return after convergence since model-based approaches face model-mismatch errors in the learning process~\citep{altman2021constrained}. Surprisingly, in Figure~\ref{fig:reward_return}, our approach consistently achieves comparable total reward returns and often surpasses other model-based and model-free methods across most of the benchmarks, except \texttt{PointPush1}. This pheromone indicates our latent model can accurately compress and reconstruct the image observation space, as shown in Figure~\ref{fig:experiemnt_setup}. In \texttt{PointPush1}, model-free methods outperform all model-based methods including ours. Our hypothesis is based on the observation in Figure~\ref{fig:experiemnt_setup}, which suggests that, in the \texttt{PointPush1}, the agent's ability to capture additional visual information is severely limited when attempting to push the yellow box, which results in inaccurate model learning. However, it's worth noting that even though all model-based methods perform similarly in \texttt{PointPush1}, our approach has fewer safety violations in Figure~\ref{fig:safety}.

% \begin{wrapfigure}{i}{0.6\textwidth}
%   \centering
%   \includegraphics[width=0.6\textwidth, height=0.38\textwidth]{fig/reward_return.pdf}
%   \caption{\textit{Reward Return}: The CPO, PPO-L, and TRPO-L are trained for 10M steps except on \texttt{DoggoGoal1} with 50M steps. The model-based Safe-SLAC and LAMBDA are trained with 1M steps except on \texttt{DoggoGoal1} with 2M environment steps. For all the benchmarks, our approach can achieve or even exceed other RL methods' performance at the end of training.}
%   \label{fig:reward_return}
% \end{wrapfigure}

\section{Conclusion}
\label{conclusion}
This paper introduces a model-based state-wise safe RL framework with pixel observations. We first learn to compress the high-dimensional image into a latent model where we establish a latent barrier-like function to encode state-wise safety constraints. We jointly conduct latent modeling, barrier-like function learning, and policy optimization to improve safety and performance simultaneously. Our approach significantly enhances safety while maintaining performance levels comparable to other established model-free and model-based safe RL methods. 
A possible future direction is to consider the distribution-based barrier function under the MDP setting for safe and unsafe states by encoding the forward invariance within the distributions, rather than the current sampled state sets.
\onecolumn
% In this study, we introduce a novel approach that combines Barrier Certificates with reinforcement learning to establish a robust safe RL framework centered around a latent model. This latent model transforms pixel observations into a lower-dimensional latent state space, enabling effective exploration in scenarios where the environment is unknown and only pixel-feedback is available. We address the challenge of ensuring safety by encoding step-wise safety constraints within a barrier function, effectively constraining the system from engaging in unsafe behaviors. Our approach significantly enhances safety while maintaining performance levels comparable to other established Safe RL methods, as demonstrated in the \textbf{Safety Gym} environment, specifically, the \texttt{SG6} benchmark. Through our approach, which involves the joint learning of the barrier function and control policies within a latent state space, our framework exhibits versatility and applicability across a wide range of high-dimensional observation-feedback exploration tasks. An intriguing avenue for future exploration involves categorizing the distribution of moving safe and unsafe states into contrastive static distributions, thereby encoding forward invariance within the distribution without a sampling approach, rather than relying on discrete sets. Additionally, the potential exists to explore alternative step-wise safety encoding methods, which could offer increased flexibility when dealing with more intricate and complex environments. 

% \acks{We thank a bunch of people.}
% \newpage
\bibliography{iclr2021_conference}

% \appendix

\end{document}